\RequirePackage{fix-cm}
\documentclass{acm_proc_article-sp}

\usepackage{graphicx}
\usepackage{mathtools}
\usepackage[noend]{algorithmic}
\usepackage{algorithm}
\usepackage{array}
\usepackage{amsmath}
\usepackage{microtype}
\usepackage{pdflscape}

\usepackage{epstopdf}
\DeclarePairedDelimiter\floor{\lfloor}{\rfloor}

\begin{document}

\title{The Great Time Series Classification Bake Off: An Experimental Evaluation of Recently Proposed Algorithms. Extended Version}

\numberofauthors{4} 

\author{
\alignauthor
Anthony Bagnall\\
       \affaddr{University of East Anglia}\\
       \affaddr{Norwich, Norfolk}\\
       \affaddr{United Kingdom}
       \email{ajb@uea.ac.uk}
\alignauthor
Aaron Bostrom\\
       \affaddr{University of East Anglia}\\
       \affaddr{Norwich, Norfolk}\\
       \affaddr{United Kingdom}
       \email{a.bostrom@uea.ac.uk}
\alignauthor
James Large\\
       \affaddr{University of East Anglia}\\
       \affaddr{Norwich, Norfolk}\\
       \affaddr{United Kingdom}
       \email{j.large@uea.ac.uk}
\and
\alignauthor
Jason Lines\\
       \affaddr{University of East Anglia}\\
       \affaddr{Norwich, Norfolk}\\
       \affaddr{United Kingdom}
       \email{j.lines@uea.ac.uk}
}

\maketitle

\begin{abstract}
In the last five years there have been a large number of new time series classification algorithms proposed in the literature. These algorithms have been evaluated on subsets of the 47 data sets in the University of California, Riverside time series classification archive. The archive has recently been expanded to 85 data sets, over half of which have been donated by researchers at the University of East Anglia. Aspects of previous evaluations have made comparisons between algorithms difficult. For example, several different programming languages have been used, experiments involved a single train/test split and some used normalised data whilst others did not. The relaunch of the archive  provides a timely opportunity to thoroughly evaluate algorithms on a larger number of datasets. We have implemented 18 recently proposed algorithms in a common Java framework and compared them against two standard benchmark classifiers (and each other) by performing 100 resampling experiments on each of the 85 datasets. We use these results to test several hypotheses relating to whether  the algorithms are significantly more accurate than the benchmarks and each other. Our results indicate that only 9 of these algorithms are significantly more accurate than both benchmarks and that one classifier, the Collective of Transformation Ensembles, is significantly more accurate than all of the others. All of our experiments and results are reproducible: we release all of our code, results and experimental details and we hope these experiments form the basis for more rigorous testing of new algorithms in the future.
\end{abstract}
\vspace{1cm}
\section{Introduction}

Time series classification (TSC) problems are differentiated from traditional classification problems because the attributes are ordered. Whether the ordering is by time or not is in fact irrelevant. The important characteristic is that there may be discriminatory features dependent on the ordering. The introduction of the UCR time series classification and clustering repository~\cite{UCRWeb} saw a rapid growth in the number of publications proposing time series classification algorithms. Prior to the summer of 2015 over 1,200 people have downloaded the UCR archive and it has been referenced several hundred times. The repository has contributed to increasing the quality of evaluation of new TSC algorithms. Most experiments involve evaluation on over forty data sets, often with correct significance testing and most authors release source code. This degree of evaluation and reproducibility is generally better than most areas of machine learning and data mining research.

However, there are still some fundamental problems with published TSC research that we aim to address. Firstly, nearly all evaluations are performed on a single train/test split. This can lead to over interpreting of results. The majority of machine learning research involves repeated resamples of the data, and we think TSC researchers should follow suit. To illustrate why, consider the following anecdote. We were recently contacted by a researcher who queried our published results for one nearest neighbour (1-NN) dynamic time warping (DTW) on the UCR train test splits. When comparing our accuracy results to theirs, they noticed that in some instances they differed by as much as 6\%. Over all the problems there was no significant difference, but clearly we were concerned, as it is a deterministic algorithm. On further investigation, we found out that our data were rounded to six decimal places, there's to eight. These differences  on single splits were caused by small data set sizes and tiny numerical differences. When resampling, there were no significant difference on individual problems when using 6 or 8 decimal places.

Secondly, there are some anomalies and discrepancies in the UCR data that can bias results. Not all of the data are normalised (e.g. Coffee) and some have been normalised incorrectly (e.g. ECG200). This can make algorithms look better than they really are. For example, most authors cite an error of 17.9\% for the Coffee data with 1-NN DTW, and most algorithms easily achieve lower error. However, 17.9\% error is for DTW on the non-normalised data. If it is normalised, 1-NN DTW has 0\% error, a somewhat harder benchmark to beat. ECG200 has been incorrectly formatted so that the sum of squares of the series can classify the data perfectly. If a classifier uses this feature it should be completely accurate. This will be a further source of bias.

Thirdly, the more frequently a fixed set of problems is used, the greater the danger of overfitting and detecting significant improvement that does not generalise to new problems. We should be constantly seeking new problems and enhancing the repository with new data. This is the only real route to detecting genuine improvement in classifier performance.

Finally, whilst the release of source code is admirable, the fact there is no common framework means it is often very hard to actually use other peoples code. We have reviewed algorithms written in C, C++, Java, Matlab, R and python. Often the code is ``research grade", i.e. designed to achieve the task with little thought to reusability or comprehensibility. There is also the tendency to not provide code that performs model selection, which can lead to suspicions that parameters were selected to minimize test error, thus biasing the results.

To address these problems we have implemented 20 different TSC algorithms in Java, integrated with the WEKA toolkit~\cite{Weka}. We have applied the following guidelines for the inclusion of an algorithm. Firstly, the algorithm must have been recently published in a high impact conference or journal. Secondly, it must have been evaluated on some subset of the UCR data. Thirdly, source code must be available. Finally, it must be feasible/within our ability to implement the algorithm in Java. This last criteria lead us to exclude at least two good candidates, described in Section~\ref{notConsidered}. Often, variants of a classifier are described within the same publication. We have limited each paper to one algorithm and taken the version we consider most representative of the key idea behind the approach.

We have conducted experiments with these algorithms and standard WEKA classifiers on 100 resamples of every data set (each of which is normalised), including the 40 new data sets we have introduced into the archive. In addition to resampling the data sets, we have also conducted extensive model selection for many of the classifiers. Full details of our experimental regime are given in Section~\ref{experimentalDesign}.

 This is one of the largest ever experimental studies conducted in machine learning. We have performed millions of experiments distributed over thousands of nodes of a large high performance computing facility. Nevertheless, the goal of the study is tightly focussed and limited. This is meant to act as a springboard for further investigation into a wide range of TSC problems we do not address. Specifically, we assume all series in a problem are equal length, real valued and have no missing values. Classification is offline, and we assume the cases are independent (i.e. we do not perform streaming classification). All series are labelled and all problems involve learning the labels of univariate time series. We are interested in testing hypotheses about the average accuracy of classifiers over a large set of problems. Algorithm efficiency and scalability are of secondary interest at this point. Detecting whether a classifier is on average more accurate than another is only part of the story. Ideally, we would like to know {\em a priori} which classifier is better for a class of problem or even be able to detect which is best for a specific data set. However, this is beyond the scope of this project.

Our findings are surprising, and a little embarrassing, for two reasons. Firstly, many of the algorithms are in fact no better than our two benchmark classifiers, 1-NN DTW and Rotation Forest. Secondly, of those 8 significantly better than both benchmarks, by far the best classifier is COTE~\cite{bagnall15cote}, an algorithm we proposed. It is on average over 8\% more accurate than either benchmark. Whilst gratifying for us, we fear that this outcome may cause some to question the validity of the study. We have made every effort to faithfully reproduce all algorithms. We have tried to reproduce published results, with varying degrees of success (as described below), and have consulted authors on the implementation where possible. Our results are reproducable, and we welcome all input on improving the code base. We must stress that COTE is by no means the final solution. All of the algorithms we describe may have utility in specific domains, and many are orders of magnitudes faster than COTE. Nevertheless, we believe that it is the responsibility of the designers of an algorithm to demonstrate its worth. We think our benchmarking results will help facilitate an improved understanding of utility of new algorithms under alternative scenarios.

All of the code is freely accessible from a repository~\cite{TSCRepo} and detailed results and data sets are available from a dedicated website~\cite{TSCWeb}.


The rest of this paper is structured as follows. In Section~\ref{algos} we review the algorithms we have implemented. In Section~\ref{experimentalDesign} we describe the data, code structure and experimental design. In Section~\ref{results} we present and analyse the results, and in Section~\ref{conc} we summarise our findings and discuss the future direction.

%
%
%

\section{Classification Algorithms}
\label{algos}

We denote a vector in bold and a matrix in capital bold. A case/instance is a pair $\{{\bf x},y\}$ with $m$ observations $x_1, \ldots, x_m$ (the time series) and discrete class variable $y$ with $c$ possible values. A list of $n$ cases with associated class labels is ${\bf T}=<{\bf X,y}>=<({\bf x_1},y_1), \ldots, ({\bf x_n},y_n)>$. A classifier is a function or mapping from the space of possible inputs to a probability distribution over the class variable values.

The large majority of time series research in the field of data mining has concentrated on alternative distance measures that can be used for clustering, query and classification. For TSC, these distance measures are almost exculsively evaluated using with a one nearest neighbour (1-NN) classifier. The standard benchmark distance measures are Euclidean distance (ED) and dynamic time warping (DTW).  Alternative techniques taken from other fields include edit distance with real penalty (ERP) and longest common subsequence (LCSS). Three more recently proposed time domain distance measures are described in Section~\ref{time}.

DTW is by far the most popular benchmark. Suppose we want to measure the distance between two series,\\ $\mathbf{a}=\{a_1,a_2,\ldots,a_m\}$ and $\mathbf{b}=\{b_1,b_2,\ldots,b_m\}$. Let $M(\mathbf{a},\mathbf{b})$ be the $m \times
m$ pointwise distance matrix between $\mathbf{a}$ and $\mathbf{b}$, where
$M_{i,j}=   (a_i-b_j)^2$. A warping path $$P=<(e_1,f_1),(e_2,f_2),\ldots,(e_s,f_s)>$$ is a set of points (i.e. pairs of indexes) that
define a traversal of matrix $M$. So, for example, the Euclidean distance $d_E(\mathbf{a,b})=\sum_{i=1}^m (a_i-b_i)^2$ is the path along the diagonal of $M$.

A valid warping path must satisfy the conditions $(e_1,f_1)=(1,1)$ and $(e_s,f_s)=(m,m)$ and
that $0 \leq e_{i+1}-e_{i} \leq 1$ and $0 \leq f_{i+1}- f_i \leq 1$
for all $i < m$.

The DTW distance between series is the path through $M$ that minimizes the total distance, subject to constraints on
the amount of warping allowed. Let $p_i=M_{a_{e_i},b_{f_i}}$ be the distance
between elements at position $e_i$ of $\mathbf{a}$ and at position $f_i$ of $\mathbf{b}$ for the $i^{th}$
pair of points in a proposed warping path $P$. The distance for any path $P$ is

\[ D_P(\mathbf{a},\mathbf{b}) =\sum_{i=1}^s p_i.\]

If $\mathcal{P}$ is the space of all possible paths, the DTW path $P^*$ is the path that has the minimum distance, i.e.
$$P^* = \min_{P \in \mathcal{P}}(D_P(\mathbf{a},\mathbf{b})).$$

The optimal path $P^*$ can be found exactly through a dynamic programming formulation. This can be a time consuming operation, and it is common to put a restriction on the amount of warping allowed. This restriction is equivalent to putting a maximum allowable distance between any pairs of indexes in a proposed path. If the warping window,  $r$, is the proportion of warping allowed, then the optimal path is constrained so that $$|e_i-f_i| \leq r\cdot m \;\;\; \forall (e_i,f_i) \in P^*.$$

It has been shown that setting $r$ through cross validation to maximize training accuracy, as proposed in~\cite{ratanamahatana05threemyths}, significantly increases accuracy~\cite{lines15elastic}.

\subsection{Time Domain Distance Based Classifiers}
\label{time}

Numerous alternatives to DTW have been proposed. In 2008, Ding {\em et al.}~\cite{ding08comparison} evaluated 8 different distance measures on 38 data sets and found none significantly better than DTW. Since then, three more elastic measures have been proposed.

{\bf Weighted DTW (WDTW)~\cite{jeong11weighted}}

Jeong {\em et al.} describe WDTW~\cite{jeong11weighted}, which adds a multiplicative weight penalty based on the warping distance between points in the warping path. It favours reduced warping, and is a smooth alternative to the cutoff point approach of using a warping window. When creating the distance matrix $M$, a weight penalty  $w_{|i-j|}$ for a warping distance of  $|i-j|$ is applied, so that

$$M_{i,j}=  w_{|i-j|} (a_i-b_j)^2.$$

A logistic weight function is used, so that a warping of $a$ places imposes a weighting of

$$w(a)=\frac{w_{max}}{1+e^{-g\cdot(a-m/2)}},$$
where $w_{max}$ is an upper bound on the weight (set to 1), $m$ is the series length and $g$ is a parameter that controls the penalty level for large warpings. The larger $g$ is, the greater the penalty for warping.

{\bf  Time Warp Edit (TWE)~\cite{marteau09stiffness}}

Marteau propose the TWE distance~\cite{marteau09stiffness}, an elastic distance metric that includes characteristics from both LCSS and DTW. It allows warping in the time axis and combines the edit distance with Lp-norms. The warping is controlled by a {\em stiffness} parameter, $\nu$. Stiffness enforces a multiplicative penalty on the distance between matched points in a manner similar to WDTW. A penalty value $\lambda$ is applied when sequences do not match.
\begin{algorithm}[!ht]
	\caption{TWE Distance(${\bf a},{\bf b}$)}
\label{algoTwe}
	\begin{algorithmic}[1]
\REQUIRE stiffness parameter $\nu$, penalty value $\lambda$
		\STATE Let $D$ be an $m+1\times m+1$ matrix initialised to zero.
		\STATE $D(1,1) \leftarrow 0$
		\STATE $D(2,1) \leftarrow {a_1}^2$
		\STATE $D(1,2)\leftarrow {b_1}^2$
		\FOR{$i \leftarrow 2$ to $m+1$}
			\STATE $D(i,1) \leftarrow D(i-1,1)+ (a_{i-2}-a_{i-1})^2$
		\ENDFOR
		\FOR{$j \leftarrow 2$ to $m+1$}
			\STATE $D(1,i) \leftarrow D(1,j-1)+ (b_{j-2}-b_{j-1})^2$
		\ENDFOR
		
		\FOR{$i \leftarrow 2$ to $m+1$}
			\FOR{$j \leftarrow 2$ to $m+1$}
			
			\IF{$i > 2$ and $j > 2$}
				\STATE $dist1 \leftarrow D(i-1,j-1)+\nu\times |i-j|\times 2 + ({a_{i-1}-b_{j-1}})^2+({a_{i-2}-b_{j-2}})^2$
			\ELSE
				\STATE $dist1 \leftarrow D(i-1,j-1)+\nu\times |i-j| + ({a_{i-1}-b_{j-1}})^2$
			\ENDIF
			\IF{$i>2$}
				\STATE $dist2 \leftarrow D(i-1,j)+(a_{i-1}-a_{i-2})^2 + \lambda+\nu$			
			\ELSE
				\STATE $dist2 \leftarrow D(i-1,j)+{a_{i-1}}^2+\lambda $			
			\ENDIF
			
			\IF{$j>2$}
				\STATE $dist3 \leftarrow D(i,j-1)+(b_{j-1}-b_{j-2})^2 + \lambda+\nu$	
			\ELSE
				\STATE $dist3 \leftarrow D(i,j-1)+{b_{j-1}}^2+\lambda $
			\ENDIF
			\STATE $D(i,j) \leftarrow $min$(dist1,dist2,dist3)$
			\ENDFOR
		\ENDFOR
		\RETURN $D(m+1,m+1)$
	\end{algorithmic}
\end{algorithm}

{\bf Move-Split-Merge (MSM)~\cite{stefan13move-split-merge}}

 Stefan {\em et al.}~\cite{stefan13move-split-merge} present MSM distance (Algorithm~\ref{algoMsm}), a metric that is conceptually similar to other edit distance-based approaches, where similarity is calculated by using a set of operations to transform a given series into a target series. Move is synonymous with a substitute operation, where one value is replaced by another. Split and merge differ from other approaches, as they attempt to add context to insertions and deletions. The split operation inserts an identical copy of a value immediately after itself, and the merge operation is used to delete a value if it directly follows an identical value.

\begin{algorithm}[!ht]
	\caption{MSM(${\bf a},{\bf b}$)}
\label{algoMsm}
	\begin{algorithmic}[1]
\REQUIRE penalty value  $c$

		\STATE Let $D$ be an $m\times m$ matrix initialised to zero.
		\STATE $D(1,1) \leftarrow |a_1-b_1|$
		\FOR{$i \leftarrow  2$ to $m$}
			\STATE $D(i,1) \leftarrow D(i-1,1)+C(a_i,a_{i-1},b_1)$
		\ENDFOR
		\FOR{$i \leftarrow  2$ to $m$}
			\STATE $D(1,i) \leftarrow D(1,i-1)+C(b_i,a_1,b+{i-1})$
		\ENDFOR
		
		\FOR{$i \leftarrow  2$ to $m$}
	       \FOR{$j \leftarrow  2$ to $m$}
			     \STATE $D(i,j) \leftarrow min(D(i-1,j-1)+|a_i-b_j|,$\\\hspace{54px}$D(i-1,j)+C(a_i,a_{i-1},b_j),$\\\hspace{54px}$D(i,j-1)+C(b_j,a_i,b_{j-1}))$
		   \ENDFOR
		\ENDFOR
		\RETURN $D(m,m)$
	\end{algorithmic}
\end{algorithm}

$$
C(a_i,a_{i-1},b_j) = \left\{ \begin{array}{l}
 	\mbox{$c$ \textbf{if} $a_{i-1} \leq a_i \leq b_j $ \textbf{or} $a_{i-1} \geq a_i \geq b_j$} \\
  	\mbox{$c+min(|a_i-a_{i-1}|,|a_i-b_j|)$ \textbf{otherwise}.}
       \end{array} \right.
$$

We have implemented WDTW, TWE, MSM and other commonly used time domain distance measures, which are available in the package weka.core.elastic\_distance\_measures. We have generated results that are not significantly different to those published when using these distances with 1-NN. In~\cite{lines15elastic} it was shown that there is no significant difference between 1-NN with DTW and with WDTW, TWE or MSM on a set of 72 problems using a single train test split. In Section~\ref{results} we revisit this result with more data and resamples rather than a train/test split.

\subsection{Differential Distance Based Classifiers}

There are a group of algorithms that are based on the first order differences of the series,
 $$a'_i = a_i-a_{i+1} \;\; i=1 \ldots m-1,$$

which we refer to as {\em diff}. Various methods that have used just the differences have been described~\cite{jeong11weighted}, but the most successful approaches combine distance in the time domain and the difference domain.

{\bf Complexity Invariant distance  (CID)~\cite{batista14cid}}

Batista {\em et al.}~\cite{batista14cid} describe a means of weighting a distance measure to compensate for differences in the complexity in the two series being compared. Any measure of complexity can be used, but Batista {\em et al.} recommend the simple expedient of using the sum of squares of the first differences (see Algorithm~\ref{algoCID}).

\begin{algorithm}[!ht]
	\caption{CID(${\bf a},{\bf b}$)}
\label{algoCID}
	\begin{algorithmic}[1]
\REQUIRE distance function $dist$
		\STATE $d \leftarrow dist({\bf a},{\bf b})$
        \STATE $c_a\leftarrow(a_1- a_2)^2$
        \STATE $c_b\leftarrow(b_1-b_2)^2$
		\FOR{$i \leftarrow  2$ to $m-1$}
			\STATE $c_a\leftarrow c_a+(a_i-a_{i+1})^2$
			\STATE $c_b \leftarrow c_b+(b_i-b_{i+1})^2$
		\ENDFOR
		\RETURN $d \cdot \frac{ \max(c_a,c_b)}{\min(c_a,c_b)} $
	\end{algorithmic}
\end{algorithm}

{\bf Derivative DTW (DD$_{DTW}$)~\cite{gorecki13derivative}}

G\'{o}recki and \L{}uczak~\cite{gorecki13derivative} describe an approach for using a weighted combination of raw series and first-order differences for NN classification with either the Euclidean distance or full-window DTW. They find the DTW distance between two series and the two differenced series. These two distances are then combined using a weighting parameter $\alpha$ (See Algorithm~\ref{dddtw}). Parameter $\alpha$ is found during training through a leave-one-out cross-validation on the training data. This search is relatively efficient as different parameter values can be assessed using pre-computed distances.

\begin{algorithm}[!ht]
	\caption{DD$_{DTW}$ (${\bf a},{\bf b}$)}
\label{dddtw}
	\begin{algorithmic}[1]
\REQUIRE weight $\alpha$, distance function $dist$, difference function {\em diff}
\STATE ${\bf c} \leftarrow$ {\em diff}$({\bf a}$)
\STATE ${\bf d} \leftarrow$ {\em diff}$({\bf b}$)
\STATE $ x \leftarrow dist({\bf a},{\bf b})$
\STATE $ y \leftarrow dist({\bf c},{\bf d})$
\STATE $ d \leftarrow \alpha \cdot x + (1-\alpha)\cdot y$
\RETURN $d$
	\end{algorithmic}
\end{algorithm}

An optimisation to reduce the search space of possible parameter values is proposed in~\cite{gorecki13derivative}. However, we could not recreate their results using this optimisation. We found that if we searched through all values of $\alpha$ in the range of $[0,1]$ in increments of 0.01, we were able to recreate the results exactly.  Testing is then performed with a 1-NN classifier using the combined distance function given in Algorithm~\ref{dddtw}.

{\bf Derivative Transform Distance (DTD$_C$)~\cite{gorecki14nonisometric}}

G\'{o}recki and \L{}uczak proposed an extension of DD$_{DTW}$ that uses DTW in conjunction with transforms and derivatives~\cite{gorecki14nonisometric}. they propose and evaluate combining DD$_{DTW}$ with distances on date transformed with the sin, cosine and Hilbert transform. We implement the cosine version (see Algorithm~\ref{dtdc}), where operation cosine transforms a series ${\bf a}$ into series ${\bf c}$ using the formula

$$c_i = \sum_{j=1}^m a_j \cos \left( \frac{\Pi}{2} \left( j - \frac{1}{2} \right)(i-1) \right)\;\; i=1 \ldots m.$$

\begin{algorithm}[!ht]
	\caption{DTDC (${\bf a},{\bf b}$)}
\label{dtdc}
	\begin{algorithmic}[1]
\REQUIRE weights $\alpha$ and $\beta$, distance function $dist$,  difference function {\em diff}

\STATE ${\bf c} \leftarrow$ {\em diff}$({\bf a}$)
\STATE ${\bf d} \leftarrow$ {\em diff}$({\bf b}$)
\STATE ${\bf e} \leftarrow$ {\em cos}$({\bf a}$)
\STATE ${\bf f} \leftarrow$ {\em cos}$({\bf b}$)

\STATE $ x \leftarrow dist({\bf a},{\bf b})$
\STATE $ y \leftarrow dist({\bf c},{\bf d})$
\STATE $ z \leftarrow dist({\bf e},{\bf f})$
\STATE $ d \leftarrow \alpha \cdot x + \beta \cdot y + (1-\alpha-\beta)\cdot z$
\RETURN $d$
	\end{algorithmic}
\end{algorithm}

DD$_{DTW}$ was evaluated on single train test splits of 20 UCR datasets, CID$_{DTW}$ on 43 datasets and DTD$_C$ on 47. We can recreate results that are not significantly different to those published for all three algorithms.

All papers claim superiority to DTW. The small sample size for DD$_{DTW}$ makes this claim debatable, but the published results for CID$_{DTW}$ and DTD$_C$ are both significantly better than DTW. On published results, DTD$_C$ is significantly more accurate than CID$_{DTW}$ and CID$_{DTW}$ is significantly better than DD$_{DTW}$. We can reproduce results not significantly different to those published for DD$_{DTW}$, CID$_{DTW}$ and DTD$_C$.

\subsection{Dictionary Based Classifiers}

Dictionary based approaches approximate and reduce the dimensionality of series by transforming them into representative words, then basing similarity on comparing the distribution of words. The core process of dictionary approaches involves forming words by passing a sliding window, length $w$, over each series, approximating each window to produce $l$ values, and then discretising these values by assigning each a symbol from an alphabet of size $\alpha$.

{\bf Bag of Patterns (BOP)~\cite{lin12bagofpatterns}}

BOP is a dictionary classifier built on the Symbolic Aggregate Approximation (SAX) method for converting series to strings~\cite{lin07sax}. SAX reduces the dimension of a series through Piecewise Aggregate Approximation (PAA)~\cite{chakrabarti02dimensionality}, then discretises the (normalised) series into bins formed from equal probability areas of the Normal distribution.

BOP works by applying SAX to each window to form a word. If consecutive windows produce identical words,  then only the first of that run is recorded. This is included to avoid over counting trivial matches. The distribution of words over a series forms a count histogram.

To classify new samples, the same transform is applied to the new series and the nearest neighbour within the training matrix found.

\begin{algorithm}[!ht]
	\caption{buildClassifierBOP(A list of $n$ cases of length $m$, ${\bf T}=\{{\bf X,y}\}$)}
	\label{bop}
	\begin{algorithmic}[1]
\REQUIRE the word length $l$, the alphabet size $\alpha$ and the window length $w$
		\STATE Let ${\bf H}$ be a list of $n$ histograms $<{\bf h}_1,\ldots,{\bf h}_n>$
		\FOR {$i \leftarrow  1$ to $n$}
			\FOR {$j \leftarrow 1$ to $m-w$}
				\STATE ${\bf q} \leftarrow x_{i,j} \ldots x_{i,j+w}$
				\STATE ${\bf r} \leftarrow $ SAX($q, l, \alpha$)
				\IF{$\neg$ trivialMatch(${\bf r, p}$)}
					\STATE $pos \leftarrow$ index(${\bf r}$) \COMMENT{{\em the function} index {\em determines the location of the word ${\bf r}$ in the count matrix ${\bf h_i}$}}
					\STATE ${h}_{i,pos} \leftarrow {\bf h}_{i,pos} + 1$
				\ENDIF
				\STATE ${\bf p} \leftarrow {\bf r}$
			\ENDFOR
		\ENDFOR
	\end{algorithmic}
\end{algorithm}

BOP sets the three parameters through cross validation. Classification of new samples is by a 1-NN with Euclidean distance between histograms as the distance measure.

{\bf Symbolic Aggregate Approximation - Vector Space Model (SAXVSM)~\cite{senin13sax_vsm}}

SAXVSM combines the SAX representation used in BOP with the vector space model commonly used in Information Retrieval. The key differences between BOP and SAXVSM is that SAXVSM forms word distributions over classes rather than series and weights these by the term frequency/inverse document frequency ($tf\cdot idf$). For SAXVSM, term frequency $tf$ refers to the number of times a word appears in a class and document frequency $df$ means the number of classes a word appears in. $tf\cdot idf$ is then defined as follows.

$\begin{displaystyle}
tfidf(tf, df) = \left.
\begin{cases}
\log{(1+tf)}\cdot \log(\frac{c}{ df }) & \text{if } df > 0 \\
0 & otherwise \\
\end{cases} \right.
\end{displaystyle}$

where $c$ is the number of classes. SAXVSM is described formally in Algorithm~\ref{saxvsm}.
\begin{algorithm}[!ht]
	\caption{buildClassifierSAXVSM(A list of $n$ cases of length $m$, ${\bf T}=\{{\bf X,y}\}$)}
	\label{saxvsm}
	\begin{algorithmic}[1]
\REQUIRE the word length $l$, the alphabet size $\alpha$ and the window length $w$
		\STATE Let ${\bf H}$ be a list of $c$ class histograms $<{\bf h}_1,\ldots,{\bf h}_c>$
		\STATE Let ${\bf M}$ be a list of $c$ class $tf\cdot idf$ $<{\bf m}_1,\ldots,{\bf m}_c>$
		\STATE Let ${\bf v}$ be a set of all SAX words found
		\FOR {$i \leftarrow  1$ to $n$}
			\FOR {$j \leftarrow 1$ to $m-w$}
				\STATE ${\bf q} \leftarrow x_{i,j} \ldots x_{i,j+w}$
				\STATE ${\bf r} \leftarrow$ SAX($q, l, \alpha$)
				\IF{$\neg$trivialMatch(${\bf r, p}$)}
					\STATE $pos \leftarrow$ index(${\bf r}$)
					\STATE $h_{y_i,pos} \leftarrow h_{y_i,pos} + 1$
					\STATE ${\bf v}$.add($r$)
				\ENDIF
				\STATE ${\bf p} \leftarrow r$
			\ENDFOR
		\ENDFOR
		\FOR {$v \in {\bf v}$}
			\STATE $pos \leftarrow$ index($v$)
			\STATE $df \leftarrow 0$
			\FOR{$i \leftarrow 1$ to $c$}
                \IF{ $h_{i,pos}>0$}
                \STATE $df \leftarrow df+1$
                \ENDIF
            \ENDFOR
			\FOR{$i \leftarrow 1$ to $c$}
				\STATE $m_{i,pos} \leftarrow tfidf(h_{i,pos}, df)$
			\ENDFOR
		\ENDFOR
	\end{algorithmic}
\end{algorithm}

Parameters $l$, $\alpha$ and $w$ are set through cross validation on the training data. Predictions are made using a 1-NN classification based on the word frequency distribution of the new case and the $tf\cdot idf$ vectors of each class. The Cosine similarity measure is used.

{\bf Bag of SFA Symbols (BOSS)~\cite{schafer15boss}}

BOSS also uses windows to form words over series, but it has several major differences to BOP and SAXVSM. Primary amongst these is that BOSS uses a truncated Discrete Fourier Transform (DFT) instead of a PAA on each window. Another difference is that the truncated series is discretised through a technique called Multiple Coefficient Binning (MCB), rather than using fixed intervals. MCB finds the disretising break points as a preprocessing step by estimating the distribution of the Fourier coefficients. This is performed by segmenting the series, performing a DFT, then finding breakpoints for each coefficient so that each bin contains the same number of elements. BOSS then involves similar stages to BOP; it windows each series to form word distribution through the application of DFT and discretisation by MCB. A bespoke distance function is used for nearest neighbour classification. This non symmetrical function only includes distances between frequencies of words that actually occur within the first histogram passed as an argument. BOSS also includes a parameter that determines whether the subseries are normalised or not.

\begin{algorithm}[!ht]
	\caption{buildClassifierBOSS(A list of $n$ cases of length $m$, ${\bf T}=\{{\bf X,y}\}$)}
	\label{boss}
	\begin{algorithmic}[1]
\REQUIRE the word length $l$, the alphabet size $\alpha$, the window length $w$, normalisation parameter $p$
		\STATE Let ${\bf H}$ be a list of $n$ histograms $<{\bf h}_1,\ldots,{\bf h}_n>$
		\STATE Let ${\bf B}$ be a matrix of $l$ by $\alpha$ breakpoints found by MCB
		\FOR {$i \leftarrow  1$ to $n$}
			\FOR {$j \leftarrow 1$ to $m-w$}
				\STATE ${\bf o}\leftarrow x_{i,j} \ldots x_{i,j+w}$
				\STATE ${\bf q} \leftarrow$ DFT($o, l, \alpha$,$p$) \COMMENT{ {\em {\bf q} is a vector of the complex DFT coefficients}}
				\STATE ${\bf q'} \leftarrow <q_1 \ldots q_{l/2}>$
				\STATE ${\bf r} \leftarrow$ SFAlookup(${\bf q', B}$)
				\IF{$\neg$trivialMatch(${\bf r, p}$)}
					\STATE $pos \leftarrow $index(${\bf r}$)
					\STATE ${h}_{i,pos} \leftarrow {h}_{i,pos} + 1$
				\ENDIF
				\STATE ${\bf p} \leftarrow {\bf r} $
			\ENDFOR
		\ENDFOR
	\end{algorithmic}
\end{algorithm}

{\bf DTW Features (DTW$_F$)~\cite{kate15features}}

Kate~\cite{kate15features} proposes a feature generation scheme that combines DTW distances to training cases and SAX histograms. A training set with $n$ cases is transformed into a set with $n$ features, where feature $x_{ij}$ is the full window DTW distance between case $i$ and case $j$. A further $n$ features are then created. These are the optimal window DTW distance between cases. Finally, SAX word frequency histograms are created for each instance using the BOP algorithm. These $a^l$ features are concatenated with the $2n$ full and optimal window DTW features. The new data set is trained with a support vector machine with a polynomial kernel with order either 1, 2 or 3, set through cross validation. DTW window size and SAX parameters are also set independently through cross validation with a 1-NN classifier. A more formal description is provided in Algorithm~\ref{dtwf}.

\begin{algorithm}[!ht]
	\caption{buildClassifierDTW$_F$(A list of $n$ cases of length $m$, ${\bf T}=\{{\bf X,y}\}$)}
	\label{dtwf}
	\begin{algorithmic}[1]
\REQUIRE the SVM order $s$, SAX word length $l$, alphabet size $\alpha$ and window length $w$, DTW window width $r$
		\STATE Let ${\bf Z}$ be a list of $n$ cases of length $2n+a^l$, ${\bf z}_1 \ldots, {\bf z}_n$ initialised to zero.
		\FOR {$i \leftarrow  1$ to $n$}
    		\FOR {$j \leftarrow  i+1$ to $n$}
                \STATE $z_{i,j} \leftarrow DTW({\bf x}_i,{\bf x}_j)$
                \STATE $z_{j,i} \leftarrow z_{i,j}$
    		\ENDFOR
 		\ENDFOR
		\FOR {$i \leftarrow  1$ to $n$}
    		\FOR {$j \leftarrow  i+1$ to $n$}
                \STATE $z_{i,n+j} \leftarrow DTW({\bf x}_i,{\bf x}_j,r)$
                \STATE $z_{n+j,i} \leftarrow z_{i,n+j}$
    		\ENDFOR
 		\ENDFOR
		\FOR {$i \leftarrow  1$ to $n$}
			\FOR {$j \leftarrow 1$ to $m-w$}
				\STATE ${\bf q} \leftarrow x_{i,j} \ldots x_{i,j+w}$
				\STATE ${\bf r} \leftarrow $ SAX($q, l, \alpha$)
				\IF{$\neg$ trivialMatch(${\bf r, p}$)}
					\STATE $pos \leftarrow$ index(${\bf r}$)
					\STATE $z_{i,2n+pos} \leftarrow z_{i,2n+pos} + 1$
				\ENDIF
				\STATE ${\bf p} \leftarrow {\bf r}$
			\ENDFOR
		\ENDFOR
        \STATE SVM.buildClassifier(${\bf Z},s)$
    \end{algorithmic}
\end{algorithm}

{\bf Published Results for Dictionary Based Classifiers}

\begin{figure}[!ht]
	\centering
       \includegraphics[width=8cm, trim={4cm 15cm 4cm 10cm}]{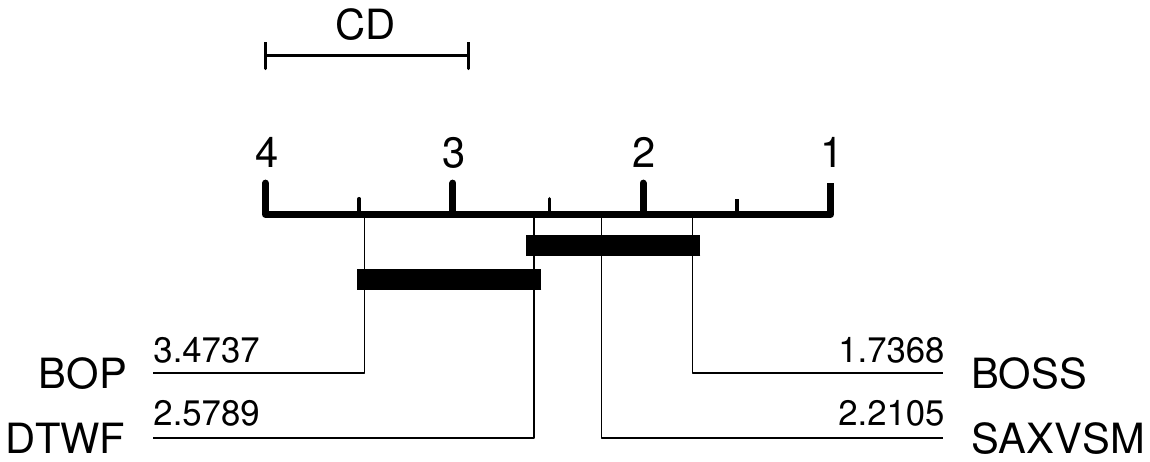}
       	\caption{Average ranks of published results on 19 data sets for BOP, SAXVSM, BOSS and DTW$_F$}
       \label{dictionary}
\end{figure}

BOP and SAXVSM were evaluated on the 20 and 19 UCR problems respectively. All algorithms used the standard single train/test split. BOSS presents results on an extended set of 58 data sets from a range of sources, DTW$_F$ uses 47 UCR data. On the 19 data sets they all have in common, BOP is significantly worse than BOSS and SAXVSM. There is no significant differencebetween DTWF, BOSS and SAXVSM (see Figure~\ref{dictionary}). Furthermore, there is no significant difference between BOSS and DTW$_F$ on the 44 datasets they have in common.

Our BOP and DTW$_F$ results are not significantly different to the published ones. We were unable to reproduce as accurate results as published for SAXVSM and BOSS. On examination of the implementation for SAXVSM provided online and by correspondence with the author, it appears the parameters for the published results were obtained through optimisation on the test data. This obviously introduces bias, as can be seen from the results for Beef. An error of 3.3\% was reported, This is far better than any other algorithm has achieved. Our results for BOSS are on average approximately 1\% worse than those published, a significant difference. Correspondence with the author and examination of the code leads us to believe this is because of a versioning problem with the code that meant the normalisation parameter was set to minimize test data error rather than train error. This would introduce significant bias.

\subsection{Shapelet Based Classifiers}
\label{shapelets}
Shapelets are time series subsequences that are discriminatory of class membership. They allow for the detection of phase-independent localised similarity between series within the same class. The original shapelets algorithm by Ye and Keogh~\cite{ye11shapelets} uses a shapelet as the splitting criterion for a decision tree. There have been three recent advances in using shapelets.

{\bf Fast Shapelets (FS)~\cite{rakthanmanon13fastshapelets}}

 Rakthanmanon and Keogh~\cite{rakthanmanon13fastshapelets} propose an extension of the decision tree shapelet approach~\cite{ye11shapelets, mueen11logicalshapelet} that speeds up shapelet discovery. Instead of a full enumerative search at each node, the fast shapelets algorithm discretises and approximates the shapelets. Specifically, for each possible shapelet length, a dictionary of SAX words is first formed. The dimensionality of the SAX dictionary is reduced through masking randomly selected letters (random projection).  Multiple random projections are performed, and a frequency count histogram is built for each class. A score for each SAX word can be calculated based on how well these frequency tables discriminate between classes. The $k$-best SAX words are selected then mapped back to the original shapelets, which are assessed using information gain in a way identical to that used in~\cite{ye11shapelets}. Algorithm~\ref{FS} gives a modular overview.

\begin{algorithm}[!ht]
	\caption{buildClassifierFS(A list of $n$ cases of length $m$, ${\bf T}=\{{\bf X,y}\}$)}
\label{FS}
\begin{algorithmic}[1]
\REQUIRE SAX the word length $l$, the alphabet size $\alpha$ and the window length $w$, number of random projections $r$, number of SAX words to convert back, $k$
\STATE Let ${\bf b}$ be an empty shapelet with zero quality
\FOR{$l \leftarrow  5$ to $m$}
	\STATE $SAXList \leftarrow$ createSaxList({\bf T}, $l$, $\alpha$,$w$)
	\STATE $SAXMap \leftarrow$ randomProjection($SAXList$,$r$)
	\STATE $ScoreList \leftarrow $scoreAllSAX($SAXList$,$SAXMap$)
	\STATE ${\bf s} \leftarrow $findBestSAX($ScoreList$, $SAXList$, $k$)
	\IF{${\bf b} < {\bf s}$}
		\STATE ${\bf b} \leftarrow {\bf s}$
	\ENDIF
\ENDFOR
\STATE $\{ {\bf T}_1, {\bf T}_2\} \leftarrow $ splitData(${\bf T},{\bf b}$)
 \IF{$\neg$ isLeaf(${\bf T}_1)$}
	\STATE buildClassifierFS(${\bf T}_1$)
\ENDIF
	
\IF{$\neg$  isLeaf(${\bf T}_2$)}
	\STATE buildClassifierFS(${\bf T}_2$)
\ENDIF	
\end{algorithmic}
\end{algorithm}

{\bf Shapelet Transform (ST)~\cite{hills14shapelet,bostrom15binary}}

Hills {\em et al.}~\cite{hills14shapelet} propose a shapelet transformation that separates the shapelet discovery from the classifier by finding the top $k$ shapelets on a single run (in contrast to the decision tree, which searches for the best shapelet at each node). The shapelets are used to transform the data, where each attribute in the new dataset represents the distance of a series to one of the shapelets. We use the most recent version of this transform~\cite{bostrom15binary} that balances the number of shapelets per class and evaluates each shapelet on how well it discriminates just one class.

\begin{algorithm}[!ht]
	\caption{BinaryShapeletSelection(A list of $n$ cases of length $m$, ${\bf T}=\{{\bf X,y}\}$)}
\label{st}
	\begin{algorithmic}[1]
\REQUIRE $min$ and $max$ length shapelet to search for the maximum number of shapelets to find $k$, number of classes $c$
		\STATE ${\bf s} \leftarrow \emptyset$
		\STATE $p \leftarrow k / c$
        \FORALL{${\bf t} \in \bf{T}$}
			\STATE $ {\bf r} \leftarrow \emptyset$
		    \FOR{$l \leftarrow  min$ to $max$}
                \STATE ${\bf W}\leftarrow$ generateCandidates(${\bf t}, l)$
			    \FORALL{subseries ${\bf a} \in {\bf W}$}
					\STATE ${\bf d} \leftarrow $ findDistances(${\bf a}, {\bf T})$
					\STATE $q \leftarrow$ assessCandidate(${\bf a, d})$
					\STATE $ {\bf r} \leftarrow {\bf r} \bigcup <{\bf a} , q>)$
				\ENDFOR
			\ENDFOR
			\STATE sortByQuality$({\bf r})$
			\STATE removeSelfSimilar$({\bf r})$
			\STATE ${\bf s} \leftarrow$ merge$({\bf s},p,{\bf r})$
		\ENDFOR
		\RETURN ${\bf s}$	
	\end{algorithmic}
\end{algorithm}

The transform described in Algorithm~\ref{st} creates a new dataset. Following~\cite{bagnall15cote,bostrom15binary} we construct a classifier from this dataset using a weighted ensemble of standard classifiers. We include $k$ Nearest Neighbour (where $k$ is set through cross validation), Naive Bayes, C4.5 decision tree, Support Vector Machines with linear and quadratic basis function kernels, Random Forest (with 500 trees), Rotation Forest (with 50 trees) and a Bayesian network. Each classifier is assigned a weight based on the cross validation training accuracy, and new data (after transformation) are classified with a weighted vote. With the exception of $k$-NN, we do not optimise parameter settings for these classifiers via cross validation.

{\bf Learned Shapelets (LS)~\cite{grabocka14learning-shapelets}}

Grabocka {\em et al.}~\cite{grabocka14learning-shapelets} describe a shapelet discovery algorithm that adopts a heuristic gradient descent shapelet search procedure rather than enumeration. LS finds $k$ shapelets that, unlike the alternatives, are not restricted to being subseries in the training data. The $k$ shapelets are initialised through a $k$-means clustering of candidates from the training data. The objective function for the optimisation process is a logistic loss function (with regularization term) $L$ based on a logistic regression model for each class. The algorithm jointly learns the weights for the regression ${\bf W}$, and the shapelets ${\bf S}$ in a two stage iterative process to produce a final logistic regression model.

\begin{algorithm}[!ht]
	\caption{learnShapelets(A list of $n$ cases of length $m$, ${\bf T}=\{{\bf X,y}\}$)}
\label{ls}
\begin{algorithmic}[1]
\REQUIRE number of shapelets $K$, minimum shapelet length $L^{min}$,  scale of shapelet length, $R$, regularization parameter,  $\lambda_W$, learning rate, $\eta$, number of iterations, $maxIter$, and softmax parameter, $\alpha$.

\STATE $ {\bf S} \leftarrow$initializeShapeletsKMeans($\bf{T}$,$K$,$R$, $L^{min}$)
\STATE $ {\bf W} \leftarrow$initializeWeights($\bf{T}$,$K$,$R$)

\FOR {$i \leftarrow 1$ to  $maxIter$}
    \STATE $ {\bf M} \leftarrow $ updateModel(${\bf T},{\bf S}, \alpha,L^{min},R$)
    \STATE $ {\bf L} \leftarrow $ updateLoss(${\bf T},{\bf M}, {\bf W}$)
    \STATE $ {\bf W}, {\bf S}\leftarrow $ updateWandS(${\bf T},  {\bf M},{\bf W},{\bf S},\eta,R,L^{min},{\bf L},\lambda_W,\alpha$)
	\IF{diverged()}
		\STATE $i = 0$
		\STATE $\eta = \eta / 3$
	\ENDIF
\ENDFOR
\end{algorithmic}
\end{algorithm}

Algorithm~\ref{ls} gives a high level view of the algorithm. LS restricts the search to shapelets of length
$\{L^{min},2L^{min},\ldots,RL^{min}\}.$ A check is performed at certain intervals as to whether divergence has occurred (line 7). This is defined as a train set error of 1 or infinite loss. The check is performed when half the number of allowed iterations is complete. This criteria meant that for some problems, LS never terminated during model selection. Hence we limited the the algorithm to a maximum of five restarts.

{\bf Published Results for Shapelet Based Classifiers}

\begin{figure}[!ht]
	\centering
       \includegraphics[width=8cm, trim={4cm 15cm 4cm 10cm}]{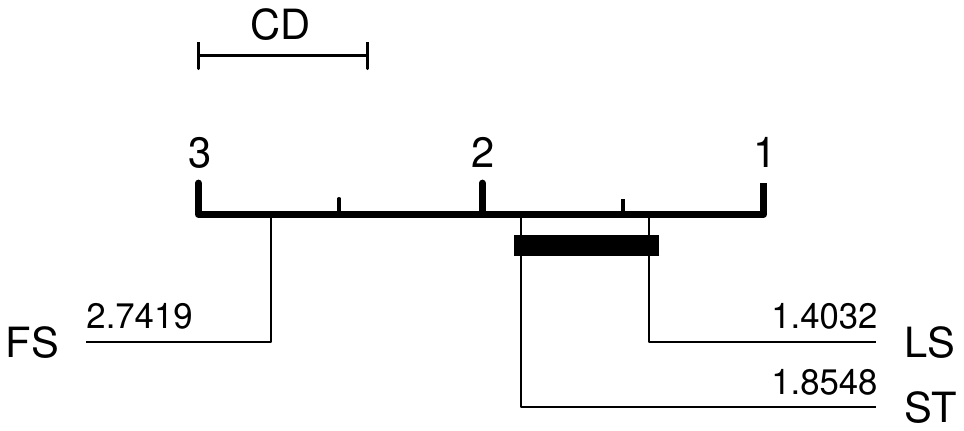}
       	\caption{Average ranks of published results for FS, LS and ST}
       \label{shapelet}
\end{figure}

FS, LS and ST were evaluated on 33, 45 and 75 data sets respectively. We can reproduce results that are not significantly different to FS and ST. The published results for FS are significantly worse than those for LS and ST (see Figure~\ref{shapelet}). There is no significant difference between the LS and ST published results.

We can reproduce the output of the code released for LS but are unable to reproduce the actual published results. The author of LS believes the difference is caused by the fact we have not included the adaptive learning rate adjustment implemented through Adagrad. We are working with him to include this enhancement.

\subsection{Interval Based Classifiers}

A family of algorithms derive features from intervals of each series. For a series of length $m$, there are $m(m-1)/2$ possible contiguous intervals. The two key decisions about using this approach are, firstly, how to deal with the huge increase in the dimension of the feature space and secondly, what to actually do with each interval. Rodriguez {\em et al.}~\cite{rodriguez05intervalbased} were the first to adopt this approach and address the first issue by using only intervals of lengths equal to powers of two and the second by calculating binary features over each intervals based on threshold rules on the interval mean and standard deviation. A support vector machine is then trained on this transformed feature set. This algorithm was a precursor to three recently proposed interval based classifiers that we have implemented.

{\bf Time Series Forest (TSF)~\cite{deng13forest}}

Deng \textit{et al.}~\cite{deng13forest} overcome the problem of the huge interval feature space by employing a random forest approach, using summary statistics (mean, standard deviation and slope) of each interval as features. Each member of the ensemble is given $\sqrt m$ intervals. A classification tree that has two bespoke characteristics is defined. Firstly, rather than evaluate all possible split points to find the best information gain, a fixed number of evaluation points is pre-defined. We assume this is an expedient to make the classifier faster, as it removes the need to sort the cases by each attribute value. Secondly, a refined splitting criteria to choose between features with equal information gain is introduced. This is defined as the distance between the splitting margin and the closest case. The intuition behind the idea is that if two splits have equal entropy gain, then the split that is furthest from the nearest case should be preferred. This measure would have no value if all possible intervals were evaluated because by definition the split points are taken as equi-distant between cases. We experimented with including these two features, but found the effect on accuracy was, if anything, negative. We found the computational overhead of evaluating all split points acceptable, hence we had no need to include the margin based tie breaker. Training a single tree involves selecting $\sqrt{m}$ random intervals, generating the mean, standard deviation and slope of the random intervals then creating and training a tree on the resulting $3\sqrt{m}$ features. Classification is by a majority vote of all the trees in the ensemble. We used the built in Weka RandomTree classifier (which is the basis for the Weka RandomForest classifier) with default parameters. This means there is no limit to the depth of the tree nor a minimum number of cases per leaf node. A more formal description is given in Algorithm~\ref{tsf}.
\begin{algorithm}[!ht]
	\caption{buildClassifierTSF(A list of $n$ cases of length $m$, ${\bf T}=\{{\bf X,y}\}$)}
\label{tsf}
	\begin{algorithmic}[1]
\REQUIRE the number of trees, $r$ and the minimum subseries length, $p$.
\STATE Let ${\bf F}=<F_1 \ldots F_r>$ be the trees in the forest.

\FOR{$i \leftarrow  1$ to $r$}
    \STATE Let ${\bf S}$ be a list of $n$ cases $<{\bf s_1}, \ldots, {\bf s_n}>$ each with $3\sqrt{m}$ attributes
    \FOR {$j \leftarrow  1$ to $\floor*{\sqrt{m}}$}
        \STATE $a=rand(1,m-p)$
        \STATE $b=rand(s+p,m)$
        \FOR {$k \leftarrow  1$ to $n$}
            \STATE $s_{k,3(j-1)+1}=$ mean(${\bf x}_k,a,b$)
            \STATE $s_{k,3(j-1)+2}=$ standardDeviation(${\bf x}_k,a,b$)
            \STATE $s_{k,3(j-1)+3}=$ slope(${\bf x}_k,a,b$)
        \ENDFOR
    \STATE $F_i.buildClassifier(\{{\bf S,y}\})$
    \ENDFOR
\ENDFOR
\end{algorithmic}
\end{algorithm}

{\bf Time Series Bag of Features (TSBF)~\cite{baygogan13tsbf}}

Time Series Bag of Features (TSBF) is an extension of TSF that has multiple stages. The first stage involves generating a subseries classification problem. The second stage forms class probability estimates for each subseries. The third stage constructs a bag of features for each original instance from these probabilities. Finally a random forest classifier is built on the bag of features representation.  Algorithm~\ref{tsbf} gives a pseudo-code overview, necessarily modularised to save space.

\begin{algorithm}[!ht]
	\caption{buildClassifierTSBF(A list of $n$ cases of length $m$, ${\bf T}=\{{\bf X,y}\}$)}
\label{tsbf}
	\begin{algorithmic}[1]
\REQUIRE the length factor $z$, the minimum interval length $a$ and the number of bins, $b$.
\STATE Let ${\bf F}$ be the first random forest and ${\bf S}$ the second.
\STATE Let $v$ be the number of intervals, $v=\left \lfloor((z \cdot m)/a) \right \rfloor$
\STATE Let $e$ be the minimum subseries length, $e=d \cdot a$
\STATE Let $w$ be the number of subseries, $w=\floor*{m/a}-d$
\STATE ${\bf S}=$generateRandomSubseries($e,w$) \COMMENT{  {\em ${\bf S}$ is the  $w \times 2$ matrix of subseries start and end points}}

\STATE ${\bf I}=$generateEqualWidthIntervals(${\bf S},d$) \COMMENT{  {\em ${\bf I}$ is the  $w \times d \times 2$ matrix of interval start and end points}}

\STATE $\{{\bf W,y'}\}=$generateIntervalFeatures(${\bf T}$,${\bf I}$)

\COMMENT{  {\em ${\bf W}$ is a set of $n\cdot w$ cases, where cases $i\cdot j$ is the summary features of intervals in the $j$th subseries of instance $i$ in training set ${\bf X}$ and $y'_{i\cdot j}$ is the class label of instance $i$.}}
\STATE ${\bf F}$.buildIncrementalClassifier($\{{\bf W,y'}\})$
\STATE ${\bf P} \leftarrow$getOOBProbabilities(${\bf F,W}$) \COMMENT{ {\em ${\bf P}$ is an $n\cdot f$ by $c$ matrix of out of bag probability estimates for the $n\cdot f$ cases in $W$.}}
\STATE ${\bf Z}\leftarrow$discretiseProbabilities(${\bf P},b$)  \COMMENT{ {\em ${\bf Z}$ is an $n\cdot w$ by $c$ matrix of integers in the range of 1 to $b$}}
\STATE ${\bf Q}\leftarrow$formHistograms(${\bf Z}$)  \COMMENT{ {\em ${\bf Q}$ is an $n$ by ($b \cdot (c-1)+c)$ list of instances where ${\bf q_i}$ corresponds to the counts the counts of the subseries derived from instance $i$ in $X$ in $Z$, split by class. Overall class probability estimates are appended to each case.}}
\STATE ${\bf S}$.buildIncrementalClassifier($\{{\bf Q,y}\})$.
\end{algorithmic}
\end{algorithm}

It can informally be summarised as follows.

\noindent {\bf Stage 1: Generate a subseries classification problem.}
\begin{enumerate}
\item Select $w$ subseries start and end points (line 7). These are the same for each of the full series. Then, for every series, repeat the following steps
\item for each of the $w$ subseries in the series, take $v$ equal width intervals (line 8) and calculate the mean, standard deviation and slope (line 9).
\item concatenate these features and the full subseries stats to form a new case with $w\cdot v+3$ attributes and class label of the original series (line 9).
\end{enumerate}

\noindent {\bf Stage 2: Produce class probability estimates for each subseries.}
\begin{enumerate}
\item Train a random forest on the new subseries dataset ${\bf W}$ (line 10). ${\bf W}$ contains $n\cdot w$ cases, each with $w\cdot v+3$ attributes. The number of trees in the random forest is determined by incrementally adding trees in groups of 50 until the out of bag error stops decreasing.
\item Find the random forest out of bag estimates of the class probabilities for each subseries (line 11).
\end{enumerate}
\noindent {\bf Stage 3: Recombine class probabilities and form a bag of patterns for each series.}
\begin{enumerate}
\item Discretise the class probability estimates for each subseries into $b$ equal width bins (line 12).
\item Bag together these discretised probabilities for each original series, ignoring the last class (line 13). If there are $c$ classes, each instance will have $w \cdot (c-1)$ attributes.
\item Add on the relative frequency of each predicted class (line 13).
\end{enumerate}
\noindent {\bf Stage 4: Build the final random forest classifier (line 14).}

 New cases are classified by following the same stages of transformation and internal classification. The number of subseries and the number of intervals are determined by a parameter, $z$. Training involves searching possible values of $z$ for the one that minimizes the out of bag error for the final classifier. Other parameters are fixed for all experiments. These are the minimum interval length (5), the number of bins for the discretisation (10), the maximum number of trees in the forest (1000), the number of trees to add at each step (50) and the number of repetitions (10).

%

{\bf Learned Pattern Similarity (LPS)~\cite{baydogan15lps}}

LPS was developed by the same research group as TSF and TSBF at Arizona State University. It is also based on intervals, but the main difference is that subseries become attributes rather than cases. Like TSBF, building the final model involves first building an internal predictive model. However, LPS creates an internal regression model rather than a classification model. The internal model is designed to detect correlations between subseries, and in this sense is an approximation of an autocorellation function. LPS selects random subseries. For each location, the subseries in the original data are concatenated to form a new attribute. The internal model selects a random attribute as the response variable then constructs a regression tree. A collection of these regression trees are processed to form a new set of instances based on the counts of the number of subseries at each leaf node of each tree. Algorithm~\ref{lps} describes the process.

\begin{algorithm}[!ht]
	\caption{buildClassifierLPS(A list of $n$ cases of length $m$, ${\bf T}=\{{\bf X,y}\}$)}
\label{lps}
	\begin{algorithmic}[1]
\REQUIRE the number of subseries, $w$ and the maximum depth of the tree, $d$.
\STATE Let $minL \leftarrow \left \lfloor (0.1\cdot m) \right \rfloor$ and $maxL \leftarrow \left \lfloor(0.9\cdot m) \right \rfloor$
\FOR{$ f \in  F$}
\STATE $e =$random($minL$,$maxL$)
\COMMENT{{\em $e$ is the subseries length}}
\STATE ${\bf A}\leftarrow$generateRandomSubseriesLocations($e$,$w$) \COMMENT{  {\em ${\bf A}$ is the  $w \times 2$ matrix of subseries start and end points}}
\STATE ${\bf B}\leftarrow$generateRandomSubseriesDifferenceLocations($e$,$w$)
\STATE ${\bf W}\leftarrow$generateSubseriesFeatures(${\bf T}$,${\bf A}$,${\bf B}$)
\COMMENT{{\em ${\bf W}$ is a set of $n \cdot e$ cases and $2w$ attributes. Attribute $i$ ($i \leq w$) is a concatenation of all of subseries with start position $A_{i,0}$ and end position $A_{i,1}$.}}
\STATE f.buildRandomRegressionTree(${\bf W}$,$d$)
\ENDFOR
\STATE Let ${\bf C}$ be a list of cases of leaf node counts ${\bf C}=<{\bf c_1},\ldots, {\bf c_n}>$
\FOR {$i=1$ to $n$}
\STATE ${\bf c_i}\leftarrow$getLeafNodeCounts(F)
\ENDFOR
\end{algorithmic}
\end{algorithm}

Less formally, LPS can be summarised as follows:

\noindent {\bf Stage 1: Construct an ensemble of $r$ regression trees.}
\begin{enumerate}
\item Randomly select a segment length $l$
\item Select $s$ segments of length $l$ from each series, transpose each segment, then concatenate. The gives a matrix ${\bf M}$ with $l\cdot n$ rows and $s$ columns.
\item Generate the difference vector for each series, transpose then concatenate. Add the new attributes to the matrix ${\bf M}$, which now has $2s$ columns.
\item Choose a random column from ${\bf M}$ as the response variable.
\item Build a random regression tree (i.e. a tree that only considers one randomly selected attribute at each level) with maximum depth of $d$.
\end{enumerate}
\noindent {\bf Stage 2: Form a count distribution over each tree's leaf node.}
\begin{enumerate}
\item For each case ${\bf x}$ in the original data, get the number of rows of ${\bf M}$ that reside in each leaf node for all cases originating from  ${\bf x}$.
\item Concatenate these counts to form a new instance. Thus if every tree had $t$ terminal nodes, the new case would have $r\cdot t$ features. In reality, each tree will have a different number of terminal nodes.
\end{enumerate}
\noindent {\bf Classification of new cases is based on a 1-nearest neighbour classification on these concatenated leaf node counts.}

There are two versions of LPS available, both of which aim to avoid the problem of generating all possible subseries. The R and C version creates the randomly selected attribute at Stage 1 on the fly at each level of the tree. This avoids the need to generate all possible subseries, but requires a bespoke tree. The second implementation (in Matlab) fixes the number of subseries to randomly select for each tree. Experiments suggest there is little difference in accuracy between the two approaches. We adopt the latter algorithm because it allows us to use the Weka RandomRegressionTree algorithm, thus simplifying the code and reducing the likelihood of bugs.

{\bf Published Results for Interval Based Classifiers}

\begin{figure}[!ht]
	\centering
       \includegraphics[width=8cm, trim={4cm 15cm 4cm 10cm}]{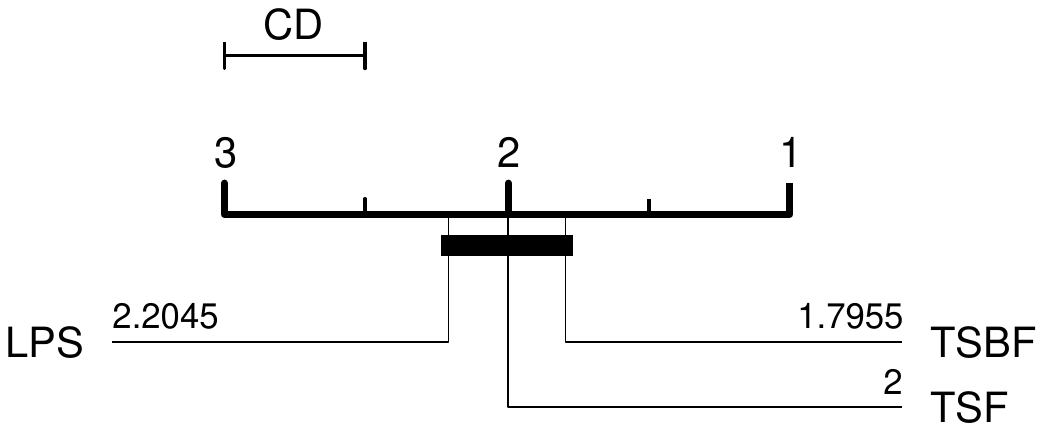}
       	\caption{Average ranks of published results for TSF, LPS and TSBF }
       \label{interval}
\end{figure}

TSF and TSBF were evaluated on the original 46 UCR problems, LPS on an extended set of 75 data sets first used in~\cite{lines15elastic} using the standard single train/test splits.
Figure~\ref{interval} shows the ranks of the published results for the problem sets they have in common. Although TSBF has the highest average rank, there is no significant difference between the classifiers at the 5\% level. Pairwise comparisons yield no significant difference between the three.

All three algorithms are stochastic, and our implementations are not identical, so there are bound to be variations between our results and those found with the original software. Our implementation of TSF has higher accuracy on 21 of the 44 datasets, worse on 23. The mean difference in accuracy is less than 1\%. There is no significant difference in means (at the 5\% level) with a rank sum test or a binomial test.

Not all of the 75 datasets LPS used are directly comparable to those in the new archive. This is because all of the new archive have been normalised, whereas many of the data proposed in~\cite{lines15elastic} are not normalised. Hence we restrict our attention to the original UCR datasets. Our LPS classifier has higher accuracy on 20 of the 44 datasets and worse on 23. The mean difference in accuracy is less than 0.02\%. Our results are not significantly different to those published when tested with a rank sum test and a binomial test. 

We can reproduce results that are not significantly different to those published for TSF and LPS. Our TSBF results are significantly worse than those published. Our TSBF classifier has higher accuracy on 9 of the 44 datasets, worse on 34. The mean difference is just over 1\%. There is no obvious reason for this discrepancy. TSBF is a complex algorithm, and it is possible there is a mistake in our implementation, but our best debugging efforts were not able to find one. It may be caused by a difference in the random forest implementations of R and Weka or by an alternative model selection method.

\subsection{Ensemble Classifiers}
\label{ensemble}

Ensembles have proved popular in recent TSC research and are highly competitive with general classification problems. TSF, TSBF and BOSS are ensembles based on the same core classifier. Other approaches, such as the $ST$ ensemble described in Section~\ref{shapelets}, use different classifier components. Two other recently proposed heterogenous TSC ensembles are as follows.

{\bf Elastic Ensemble (EE)~\cite{lines15elastic}}

 The EE is a combination of nearest neighbour (NN) classifiers that use elastic distance measures. Lines and Bagnall~\cite{lines15elastic} show that none of the individual components of EE significantly outperforms DTWCV. However, we demonstrate that by combining the predictions of 1-NN classifiers built with these distance measures and using a voting scheme that weights according to cross-validation training set accuracy, we can significantly outperform DTWCV.
  The 11 classifiers in EE are 1-NN with Euclidean distance (ED), full dynamic time warping (DTW), DTW with window size set through cross validation (DTWCV), derivative DTW with full window and window set through cross validation (DDTW and DDTWCV), weighted DTW (WDTW) and derivative weighted DTW (WDDTW)~\cite{jeong11weighted}, longest common subsequence (LCSS), Edit Distance with Real Penalty (ERP), Time Warp Edit (TWE) distance~\cite{marteau09stiffness}, and the Move-Split-Merge (MSM) distance metric~\cite{stefan13move-split-merge}.

{\bf Collective of Transformation Ensembles (COTE)~\cite{bagnall15cote}}

Bagnall {et al.} propose the meta ensemble COTE, a combination of classifiers in the time, autocorrelation, power spectrum and shapelet domain. The components of EE and ST are pooled with classifiers built on a version of autocorrelation transform (ACF) and power spectrum (PS) transform. EE uses the 11 classifiers described above. ACF and PS employ the same 8 classifiers used in conjunction with the shapelet transform. We use the classifier called flat-COTE in~\cite{bagnall15cote}. This involves pooling all 35 classifiers into a single ensemble with votes weighted by train set cross validation accuracy.

\subsection{Summary}
\label{notConsidered}

We have grouped the algorithms for clarity, but the classifications are overlapping. For example, TSBF is an interval based and ensemble based approach and LPS is based on auto-correlation. Table~\ref{summaryType} gives the break down of algorithm verses approach.

\begin{table}
\begin{center}
\caption{A summary of algorithms and the component approaches underlying them. Approaches are nearest neighbour classification (NN), time domain distance function (time), derivative based distance function (deri), shapelet based (shpt), interval based (int), dictionary based (dict), auto-correlation based (auto) and ensemble (ens)  }
\label{summaryType}
\scriptsize
\begin{tabular}{|c|c|c|c|c|c|c|c|c|} \hline
        & NN & time  & deri & shpt & int & dict & auto & ens \\ \hline
        & NN & time  & deri & shpt & int & dict & auto & ens \\ \hline
WDTW    & x & x &   &   &   &   &   &\\
TWE     & x & x &   &   &   &   &   &\\
MSM     & x & x &   &   &   &   &   &\\
CID     & x & x & x &   &   &   &   &\\
DD$_{DTW}$  & x & x & x &   &   &   & &  \\
DTD$_C$ & x & x & x &   &   &   &   &\\
ST      &   &   &   & x &   &   &   & x\\
LS      &   &   &   & x &   &   &   &\\
FS      &   &   &   & x &   &   &   &\\
TSF     &   &   &   &   & x &   &   &x\\
TSBF    &   &   &   &   & x &   &   &x\\
LPS     & x &   & x &   & x &   & x & \\
BOP     & x &   &   &   &   & x &   & \\
SAXVSM  & x &   &   &   &   & x &   & \\
BOSS    & x &   &   &   &   & x & x  & x\\
DTW$_F$ & x & x&   &   &  & x &    & x \\
EE      & x & x & x  &   &   &  &   & x \\
COTE    & x & x & x  & x &   &  & x & x \\  \hline
\end{tabular}
\end{center}
\end{table}

There are many other approaches that have been proposed that we have not included due to time constraints and failure to meet our inclusion criteria. Two worthy of mention are Silva {\em et al.}'s Recurrence Plot Compression Distance (RPCD)~\cite{silva13recurrence} and Fulcher and Jones's feature-based linear classifier (FBL)~\cite{fulcher14comparative}. RPCD involves trsansforming each series into a 2 dimensional recurrence plot then measuring similarity based on the size of the MPEG1 encoding of the concatenation of the resulting images. We were unable to find a working Java based MPEG1 encoder, and the technique seems not to work with the MPEG4 encoders we tried. FBL involves generating a huge number of possible features which are filtered with a forward selection mechanism for a linear classifier. The technique utilises built in matlab functions to generate thousands of features. Unfortunately these functions are not readily available in Java, and we considered it infeasible to attempt such as colossal task. It is further worth noting that COTE produces significantly better results than both RPCD and FBL~\cite{bagnall15cote}.

\section{Data and Experimental Design}
\label{experimentalDesign}

The 85 datasets are described in detail on the website~\cite{TSCWeb}. The collection is varied in terms of data characteristics: the length of the series ranges from 24 (ItalyPowerDemand) to 2709 (HandOutlines); train set sizes vary from 16 to 8926; and the number of classes is between 2 and 60.  The data are from a wide range of domains, with an over representation of image outline classification problems. We are introducing four new food spectra data sets Ham, Meat, Strawberry and Wine. These were all created by the Institute of Food Research, part of the Norwich Research Park, as were the three spectra data already in the UCR (Beef, Coffee and OliveOil). Table~\ref{type} gives the breakdown of number of problems per category.

\begin{table}
\begin{center}
\caption{Number of datasets by problem type}
\label{type}
\begin{tabular}{|c|c|} \hline
Image Outline   &   29\\
Sensor Readings &   16\\
Motion Capture	&    14\\
Spectrographs	& 7   \\
ECG measurements&	7\\
Electric Devices&	6\\
Simulated	    & 6 \\ \hline
Total           & 85 \\ \hline
\end{tabular}
\end{center}
\end{table}

We run the same 100 resample folds on each problem for every classifier. The first fold is always the original train test split. The other resamples are stratified to retain class distribution in the original data sets. These resample datasets can be exactly reproduced.

Each classifier must be evaluated 8,500 times. Model selection is repeated on every training set fold. We used the parameter values searched in the relevant publication as closely as possible. The parameter values we search are listed in Table~\ref{parameters}. We allow each classifier a maximum 100 parameter values, each of which we assess through a cross validation on the training data. The number of cross validation folds is dependent on the algorithm. This is because the overhead of the internal cross validation differs. For the distance based measures it is as fast to do a leave-one-out cross validation as any other.  For others we need a new model for each set of parameter values. This means we need to construct 850,000 models for each classifier. When we include repetitions caused by bugs, we estimate we have conducted over 30 million distinct experiments over six months.

The datasets vary greatly in size. The eight largest (grouped by the working title `the pigs') are ElectricalDevices, FordA, FordB, HandOutlines, NonInvasive1, NonInvasive2, StarlightCurves and UWaveGestureLibraryAll. We have had to sub-sample these data sets for the model selection stages, in particular for the slower algorithms such as ST, LPS and BOSS. Full details of the sampling performed are in the code documentation.

\begin{table*}
\begin{center}
\caption{Parameter settings and ranges for TSC algorithms. The notation is overloaded in order to maintain consistency with authors' original parameter names. }
\label{parameters}
\begin{tabular}{|c|l|l|} \hline
            & Parameters & CV Folds \\ \hline
WDTW        & $g \in \{0,0.01,\ldots,1\}$ & LOOCV\\
TWE         &$\nu \in \{0.00001, 0.0001, 0.001 ,0.01, 0.1, 1\} $ and $\lambda \in \{0, 0.25, 0.5, 0.75, 1.0\}$    & LOOCV\\
MSM         &$c \in \{0.01, 0.1, 1, 10, 100\}$ & LOOCV\\
CID         & $r \in \{0.01, 0.1, 1, 10, 100\}$    & LOOCV  \\
DD$_{DTW}$  & $a \in \{0,0.01,\ldots,1\}$  & LOOCV\\
DTD$_C$     & $ a \in \{0,0.1,\ldots,1\}$, $b \in \{0,0.1,\ldots,1\}$ & LOOCV\\
ST          & min=3, max=m-1 $k=10n $ & 0\\
LS          & $\lambda \in \{0.01,0.1\}$, $L \in \{0.1,0.2\}$, $R \in \{2,3\}$  & 3 \\
FS          & $r=10$, $k=10$, $l= 16$ and $\alpha = 4$ & 0 \\
TSF         & $r=500$ &   0   \\
TSBF     &$z \in\{0.1,0.25,0.5,0.75\}$, $a=5$, $b=10$ & LOOCV\\
LPS      & $w=20$, $d \in{2,4,6}$, & LOOCV \\
BOP      & $ \alpha \in {2, 4, 6, 8}$, w from 10\% to 36\% of $m$, $l \in {2^i | i=1 to \log(w/2)}$ & LOOCV \\
SAXVSM   & $ \alpha \in {2, 4, 6, 8}$, w from 10\% to 36\% of $m$, $l \in {2, 4, 6, 8}$ & LOOCV  \\
BOSS     & $ \alpha =4$, $w$ from 10 to $m$, with $\min(200,\sqrt(m))$ values,  $l \in {8, 10, 12, 14, 16}$ & LOOCV\\
DTW$_F$  & DTW parameters 0 to 0.99, SAX parameters as per BOP, SVM kernel degree $\{1,2,3\}$ & 10 \\
EE       & constituent classifier parameters only & 0  \\
COTE     & constituent classifier parameters only& 0 \\  \hline
\end{tabular}
\end{center}
\end{table*}

We follow the basic methodology described in~\cite{demsar06comparisons} when testing for significant difference between classifiers. For any single problem we can compare differences between two or more classifiers over the 100 resamples using standard parametric tests (t-test for two classifiers, ANOVA for multiple classifiers) or non parametric test (binomial test or the Wilcoxon sign rank test for two classifiers, Friedman test for multiple classifiers). However, the fact we are resampling data means the observations are not independent and we should be careful interpreting to much into the results for a single problem. The real benefit of resampling is to reduce the risk of bias introduced through overfitting on a single sample. Our main focus of interest is relative performance over multiple data sets. Hence, we average accuracies over all 100 resamples, then compare classifiers by ranks using the Friedman test and a {\em post-hoc} pairwise Nemenyi test to discover where the differences lie.

\section{Results}
\label{results}
Due to space constraints, we present an analysis of our results rather than presenting the full data. All of our results and spreadsheets to derive the graphs are available from~\cite{TSCWeb}.

\subsection{Benchmark Classifiers}

We believe that newly proposed algorithms should add some value in terms of accuracy or efficiency over sensible standard approaches which are generally much simpler and better understood. The most obvious starting point for any classification problem is to use a standard classifier that treats each series a vector (i.e. make no explicit use of any autocorellation structure). Some characteristics that make TSC problems hard include having few cases, long series (large number of attributes) many of which are redundant or correlated. These are problems that are well studied in machine learning and classifiers have been designed to compensate for them. TSC characteristics that will confound traditional classifiers include discriminatory features in the autocorrelation function, phase independence within a class and imbedded discriminatory subseries. However, not all problems will have this characteristic, and benchmarking against standard classifiers may give insights into the problem characteristics. We have experimented with Weka versions of C4.5  (C45), naive Bayes (NB), logistic Regression (logistic), support vector machine with linear (SVML) and quadratic kernel (SVMQ), multilayer perceptron (MLP), random forest (with 500 trees) (RandF) and rotation forest (with 50 trees) (RotF). In TSC specific research, the starting point with most investigations is 1-NN with Euclidean distance (ED). This basic classifier is a very low benchmark for comparison and is easily beaten with other standard classifiers. A more useful benchmark is 1-NN dynamic time warping with a warping window set through cross validation (DTW)~\cite{ratanamahatana05threemyths}.

\begin{figure}[!ht]
	\centering
       \includegraphics[width=8cm, trim={4cm 10cm 4cm 10cm},clip]{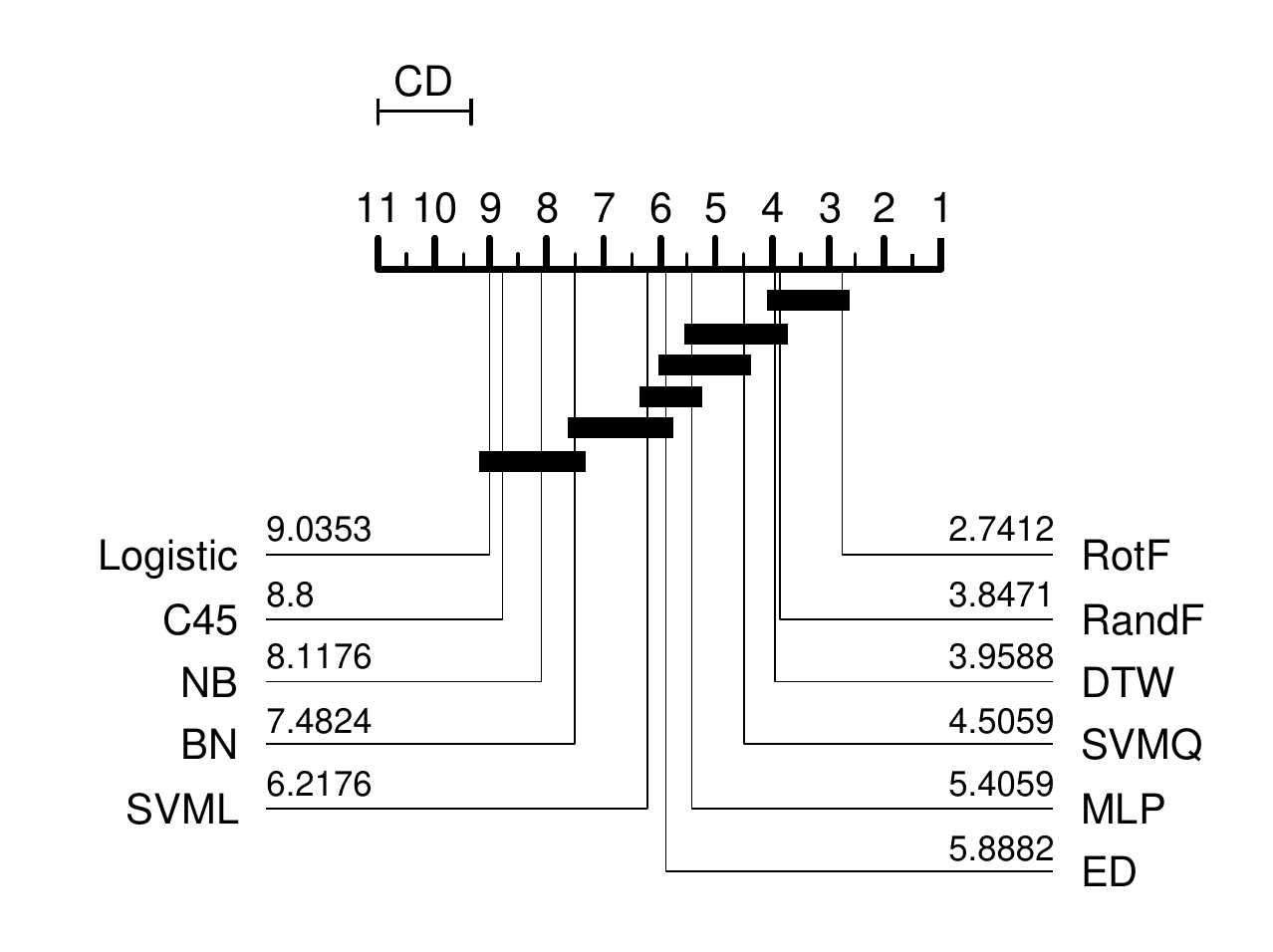}
       	\caption{Critical difference diagram for 11 potential benchmark classifiers.}
       \label{benchmarks}
\end{figure}

RotF, RandF and DTW form a clique of classifiers better than the others. Based on these results, we select RotF and DTW as our two benchmarks classifiers. Head to head, RotF has significantly better accuracy on 43 problems, DTW on 33, and no difference on 9 data sets.

\subsection{Comparison Against Benchmark Classifiers}

Table~\ref{summaryVsDTW} shows the summary of the pairwise results of the 19 classifiers against DTW and RotF. Nine classifiers are significantly better than both benchmarks: COTE; ST; BOSS; EE; DTW$_F$; TSF; TSBF; LPS; and MSM. BOP, SAXVSM and FS are all significantly worse than both the benchmarks. This reflects the published FS results, but is worse than expected for BOP and SAXVSM.

\begin{table*}
\begin{center}
\caption{A summary of algorithm performance grouped based on significant difference to DTW andRotF. The column prop gives the proportion of problems where the classifier has a significantly higher mean accuracy over 100 resamples. The column mean gives the mean difference in mean accuracy over all 85 problems.}
\label{summaryVsDTW}
\begin{tabular}{|c|c|c||c|c|c|} \hline
Classifier  &   Prop better   & Mean difference & Classifier  &   Prop better   & Mean difference \\ \hline
\hline \multicolumn{3}{|c||}{Significantly better than DTW} & \multicolumn{3}{c|}{Significantly better than RotF}\\ \hline
COTE        &  96.47\% &   8.12\% & COTE    &   84.71\% & 8.14\% \\
EE          &  95.29\% &   3.51\% & ST      &   75.29\% & 6.15\% \\
BOSS        &  82.35\% &   5.76\% & BOSS    &   63.53\% & 5.78\% \\
ST          &  80.00\% &   6.13\% & TSF     &   63.53\% & 1.93\% \\
DTW$_F$     &  75.29\% &   2.87\% & LPS     &   60.00\% & 1.86\% \\
TSF         &  68.24\% &   1.91\% & EE      &   58.82\% & 3.54\% \\
TSBF        &  65.88\% &   2.19\% & DTW$_F$ &   58.82\% & 2.89\% \\
MSM         &  62.35\% &   1.89\% & MSM     &   57.65\% & 1.91\% \\
LPS         &  61.18\% &   1.83\% & TSBF    &   56.47\% & 2.22\% \\ \cline{4-6}
WDTW        &  60.00\% &   0.20\% & \multicolumn{3}{c|}{Not significantly different to RotF}\\ \cline{4-6}
DTD$_C$     &  52.94\% &   0.79\% & CID$_{DTW}$&48.24\% & 0.56\%   \\
CID$_{DTW}$ &  50.59\% &   0.54\% & DTD$_C$   & 47.06\% & 0.82\% \\ \cline{1-3}
 \multicolumn{3}{|c||}{Not significantly different to DTW}  & DD$_{DTW}$& 45.88\% & 0.44\%   \\ \cline{1-3}
DD$_{DTW}$  &  56.47\% &  0.42\%  &TWE       & 45.88\% & 0.40\% \\
RotF        &  56.47\% &  -0.02\% &WDTW      & 44.71\% & 0.22\% \\
TWE         &  49.41\% &  0.37\%  &LS        & 44.71\% & -2.97\%\\ \cline{1-3}
\multicolumn{3}{|c||}{Significantly worse than DTW} & DTW       & 43.53\% & 0.02\% \\   \hline
LS         & 47.06\% &  -2.99\%  & \multicolumn{3}{c|}{Significantly worse than RotF} \\ \cline{4-6}
SAXVSM     & 41.18\% &  -3.29\% &  BOP       & 34.12\% &-3.03\%\\
BOP        & 37.65\% &  -3.05\% &  SAXVSM    & 31.76\% &-3.26\%\\
FS         & 30.59\% &  -7.40\% &  FS        & 22.35\% &-7.38\% \\ \hline

\hline
\end{tabular}
\end{center}
\end{table*}

\subsection{Comparison of All TSC Algorithms}

\begin{figure}[!ht]
	\centering
       \includegraphics[width=8cm, trim={4cm 10cm 4cm 10cm},clip]{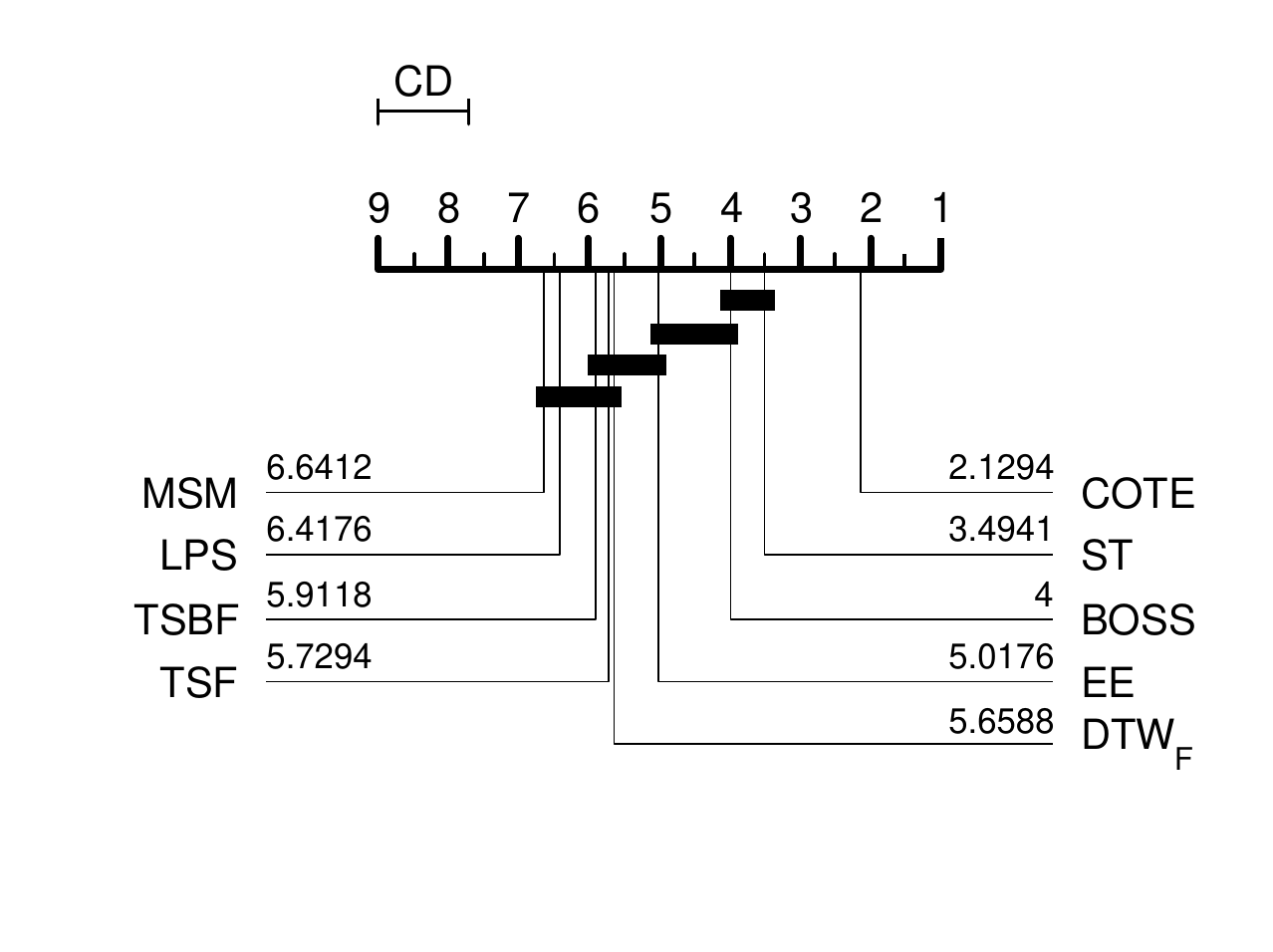}
       	\caption{Critical difference diagram for the 9 classifiers significantly better than both benchmark classifiers.}
       \label{all}
\end{figure}

Figure~\ref{all} shows the critical difference for the nine classifiers that are significantly better than both benchmarks. The most obivous conclusion from this graph is that COTE is significantly better than the others. EE and ST are components of COTE, hence this result demonstrates the benefits of combining classifiers on alternative feature spaces. The second distinguishing feature is the good performance of BOSS, and to a lesser degree, DTW$_F$. We discuss these results in detail below.

\subsection{Results by Algorithm Type}
\vspace{0.5cm}

{\bf Time Domain Distance Based Classifiers.}
Of the three distance based approaches we evaluated (TWE, WDTW and MSM), MSM is the highest rank (9$^{th}$) and is the only one significantly better than both benchmarks. WDTW (ranked 14$^{th}$) is better than DTW but not RotF. This conclusion contradicts the results in~\cite{lines15elastic} which found no difference between all the elastic algorithms and DTW. This demonstrates that whilst there is a significant improvement, the benefit is small. MSM is under 2\% on average better than DTW and RotF. The average of average differences in accuracy between WDTW and DTW is only 0.2\%. The fact we are resampling has allowed us to detect such as small improvement. We made no attempts to speed up the distance measures, and of all the measures used in~\cite{lines15elastic}, MSM and TWE were by far the slowest. These results indicate it may be worthwhile examining speed ups for MSM.

 {\bf Difference Based Classifiers.}
 In line with published results, two of the difference based classifiers, CID$_{DTW}$  and DTD$_C$ are significantly better than DTW, but the mean improvement is very small (under 1\%). None of the three approaches are significantly different to RotF. We believe this highlights an over-reliance on DTW as a benchmark. In line with the original description we set the CID$_{DTW}$ as the optimal for DTW. Setting the window to optimise the CID$_{DTW}$ distance instead might well improve performance.

{\bf Dictionary Based Classifiers.} The results for window based dictionary classifiers are confusing.
SAXVSM and BOP are significantly worse than the benchmarks and ranked 18$^{th}$ and 19$^{th}$ overall respectively. This would seem to suggest there is little merit in dictionary transformations for TSC. However, the BOSS ensemble is one the most accurate classifiers we tested (ranked 3$^{rd}$). It is significantly better than both benchmarks and is ranked third overall. The main differences between BOP and BOSS are that BOSS uses a Fourier transformation rather than PAA and employs a data driven discretisation rather than arbitrary break points in the Normal distribution. This indicates that there may be further scope for window based spectral classifiers. The use of an ensemble also significantly improves accuracy. It would be interesting to detect which difference contributes most to the improved performance of the BOSS ensemble. DTW$_F$ also did well (ranked 5$^{th}$). Including SAX features significantly improves  DTW$_F$, so our conjecture is that the DTW features are compensating for the datasets that BOP does poorly on, whilst gaining from those it does well at. This would support the COTE argument for combining features from different representations.

{\bf Shapelet Based Classifiers.}
FS is the least accurate classifier we tested and is significantly worse than the benchmarks. LS is not significantly better than either benchmark and in fact it is significantly worse than DTW. Our FS algorithm reproduces published results and we believe is faithful to the original. We have put considerable effort into debugging LS and have been in correspondence with the author. He believes the difference is caused by the fact we have not included the adaptive learning rate adjustment implemented through Adagrad. We are working with him to include this enhancement. Conversely, the ST has exceeded our expectations. It is significantly better than both benchmarks is the second most accurate classifier overall, significantly better than six of the other eight classifiers that beat both benchmarks. The changes proposed in~\cite{bostrom15binary} have not only made it much faster, but have also increased accuracy. Primary amongst these changes is balancing the number of shapelets per class and using a one-vs-many shapelet quality measure. However, ST is the slowest of all the algorithms we assessed and there is scope to increase the speed without compromising accuracy.

{\bf Interval Based Classifiers.} The interval based approaches, TSF, TSBF and LPS, are all significantly better than both the benchmarks. This gives clear support to the idea of interval based approaches. There is no significant difference between them. Hence, based on this evidence, we conclude there is definite value in interval based algorithms and would favour TSF for its simplicity.

{\bf Ensemble Classifiers.} The top seven classifiers are all ensembles. This is strong evidence to support the view that ensembling is one of the simplest ways of improving a classifier. It seems highly likely the other classifiers would benefit from a similar approach. One of the key ensemble design decisions is promoting diversity without compromising accuracy. TSF, TSBF and LPS do this through the standard approach of sampling the attribute space. BOSS ensembles identical classifiers with different parameter settings. ST and EE engender diversity though classifier heterogeneity. Employing different base classifiers in an ensemble is relatively unusual, and these results would suggest that it might be employed more often. COTE is significantly better than all other classifiers. It promotes diversity through employing different transformations/data representations and weighting by a training set accuracy estimate. Its simplicity is its strength. These experiments suggest COTE may be even more accurate if it were to assimilate BOSS and an interval based approach.

\subsection{Results by Problem Type}

Table~\ref{byProblem} shows the performance of algorithms against problem type. The data is meant to give an indication as to which family of approaches may be best for each problem type. The sample sizes are small, so we must be careful drawing too many conclusions. However, this table does indicate how evaluation can give insights into problem domains. So, for example, Shapelets are best on 4 out of 6 of the ElectricDevice problems and 3 out of 6 ECG datasets, but only 26\% of problems overall. This makes sense in terms of the applications, because the profile of electricity usage and ECG irregularity will be a subseries of the whole and largely phase independent. Vector classifiers are best on 43\% of the Spectrograph data sets. COTE is the best algorithm on over 40\% of the image outline problems. This suggests that there are a range of features that help classify these problems and no one representation is likely to be sufficient.

\begin{table*}
\begin{center}
\caption{Best performing algorithms split by problem type. Each entry is the percentage of problems of that type a member of a class of algorithm is most accurate for. }
\label{byProblem}
\begin{tabular}{|c|c|c|c|c|c|c|c|c|} \hline
Problem         &	COTE  &	Dictionary & Difference &	Elastic	& Interval &	Shapelet &	Vector& Counts	\\ \hline
Image Outline   &24.14\% &	13.79\% & 6.90\% &	17.24\% &	 0.00\% &	17.24\% &	20.69\% & 29 \\
Sensor Readings &38.89\% &	 0.00\% & 5.56\% &	11.11\% &	 5.56\% &	22.22\% &	16.67\% & 18\\
Motion Capture   &35.71\% &	21.43\% & 7.14\% &	 7.14\% &	14.29\% &	14.29\% &	 0.00\% & 14 \\
Spectrographs    & 0.00\% &	 0.00\% & 0.00\% &	 0.00\% &	 0.00\% &	 0.00\% &	100.00\% & 7\\
Electric Devices& 0.00\% &	16.67\% & 0.00\% &	 0.00\% &	16.67\% &	66.67\% &	 0.00\% & 6 \\
ECG measurements&33.33\% &	16.67\% & 0.00\% &	 0.00\% &	 0.00\% &	50.00\% &	 0.00\% & 6 \\
Simulated       &40.00\% &	20.00\% & 0.00\% &	20.00\% &	 0.00\% &	20.00\% &	 0.00\% & 5\\
Overall         &27.06\% &	11.76\% &	4.71\% &	10.59\% &	 4.71\% &	22.35\% &	 18.82\% &\\
Counts          &	23	 &   10	     & 4	    & 9	        & 4	        & 19	    &  16   & 85 \\ \hline
\end{tabular}
\end{center}
\end{table*}

\section{Conclusions}
\label{conc}
The primary goal of this series of benchmark experiments is to promote reproducible research and provide a common framework for future work in this area.

We view data mining as a practical area of research, and our central motivation is to find techniques that work. Received wisdom is that DTW is hard to beat. Our results confirm this to a degree (7 out of 19 algorithms fail to do so), but recent advances show it not impossible.

Overall, our results indicate that COTE is, on average, clearly superior to other published techniques. It is on average 8\% more accurate than DTW. However, COTE is a starting point rather than a final solution. Firstly, the no free lunch theorem leads us to believe that no classifier will dominate all others. The research issues of most interest are what types of algorithm work best on what types of problem and can we tell {\em a priori} which algorithm will be best for a specific problem. Secondly, COTE is hugely computationally intensive. It is trivial to parallelise, but its run time complexity is bounded by the Shapelet Transform, which is $O(n^2m^4)$ and the parameter searches for the elastic distance measures, some of which are $O(n^3)$. An algorithm that is faster than COTE but not significantly less accurate would be a genuine advance in the field. Finally, we are only looking at a very restricted type of problem. We have not considered multi-dimensional, streaming, windowed, long series or semi-supervised TSC, to name but a few variants. Each of these subproblems would benefit from a comprehensive experimental analysis of recently proposed techniques.

We are constantly looking for new areas of application and we will include any new data sets that are donated in an ongoing evaluation. We will happily evaluate anyone else's algorithm if it is implemented as a WEKA classifier (with all model selection performed in the method buildClassifier) and if it is computationally feasible. If we are given permission we will release any results we can verify through the associated website.

For those looking to build a predictive model for a new problem we would recommend starting with DTW, RandF and RotF as a basic sanity check and benchmark. We have made little effort to perform model selection for the forest approaches because it is generally accepted they are robust to parameter settings, but some consideration of forest size and tree parameters may yield improvements. However, our conclusion is that using COTE will probably give you the most accurate model. If a simpler approach is needed and the discriminatory features are likely to be embedded in subseries, then we would recommend using TSF or ST if the features are in the time domain (depending on whether they are phase dependent or not) or BOSS if they are in the frequency domain. If a whole series elastic measure seems appropriate, then using EE is likely to lead to better predictions than using just DTW.

Finally, we stress that accuracy is not the only consideration when assessing a TSC algorithm. Time and space efficiency are often of equal or greater concern. However, if the only metric used to support a new TSC is accuracy on these test problems, then we believe that evaluation should be transparent and comparable to the results we have made available. If a proposed algorithm is not more accurate than those we have evaluated, then some other case for the algorithm must be made.

\section*{Acknowledgment}

This work is supported by the UK Engineering and Physical Sciences Research Council (EPSRC)  [grant number EP/ M015087/1]. The experiments were carried out on the High Performance Computing Cluster supported by the Research and Specialist Computing Support service at the University of East Anglia. We would particularly like to thank Leo Earl for his help and forbearance with our unreasonable computing requirements.

\bibliographystyle{plain}
\bibliography{bakeoff}

\end{document}